\documentclass{article} % For LaTeX2e
\usepackage{iclr2026_conference,times}

\usepackage{amsmath,amsfonts,bm}

\def\eqref#1{equation~\ref{#1}}
\def\1{\bm{1}}

\DeclareMathAlphabet{\mathsfit}{\encodingdefault}{\sfdefault}{m}{sl}
\SetMathAlphabet{\mathsfit}{bold}{\encodingdefault}{\sfdefault}{bx}{n}

\usepackage{hyperref}
\usepackage{url}
\usepackage{graphicx}
\usepackage{booktabs}    
\usepackage{longtable}                      \usepackage{subcaption}   
\usepackage[capitalize,noabbrev]{cleveref}
\usepackage{tcolorbox}
\tcbuselibrary{listings,breakable}
\usepackage{comment}

\usepackage{multirow}                                   
\usepackage{array}       % Extended column definitions        
\usepackage{tabularx}    % Tables with auto-width columns                         
\title{One Size Fits None:\\ Heuristic Collapse in LLM Investment Advice}

\author{Jillian Ross \\
Computer Science and Artificial Intelligence Laboratory \\
Massachusetts Institute of Technology \\
Cambridge, MA, USA \\
\texttt{jillianr@mit.edu} \\
\And
Andrew W. Lo \\
Sloan School of Management \\
Massachusetts Institute of Technology \\
Cambridge, MA, USA \\
\texttt{alo-admin@mit.edu} \\
}

\iclrfinalcopy
\begin{document}

\maketitle

\begin{abstract}
Large language models are increasingly deployed as advisors in high-stakes domains --- answering medical questions, interpreting legal documents, recommending financial products --- where good advice requires integrating a user's full context rather than responding to salient surface features. We investigate whether frontier LLMs actually do this, or whether they instead exhibit heuristic collapse: a systematic reduction of complex, multi-factor decisions to a small number of dominant inputs. We study the phenomenon in investment advice, where legal standards explicitly require individualized reasoning over a client's full circumstances. Applying interpretable surrogate models to LLM outputs, we find systematic heuristic collapse: investment allocation decisions are largely determined by self-reported risk tolerance, while other relevant factors contribute minimally. We further find that web search partially attenuates heuristic collapse but does not resolve it. These findings suggest that heuristic collapse is not resolved by web search augmentation or model scale alone, and that deploying LLMs as advisors requires auditing input sensitivity, not just output quality.
\end{abstract}

\section{Introduction}

Large language models (LLMs) process the context they are given, but there is no guarantee they reason over all of it. In high-stakes advisory settings — financial planning, medical triage, legal interpretation — this distinction matters enormously. A model that acknowledges a user's full profile but effectively conditions its output on one or two salient features is not providing personalized advice. It is merely providing the appearance of it.

We call this phenomenon \textit{heuristic collapse}: the reduction of a complex, multi-factor decision to dependence on a small number of salient features, which produces outputs that are superficially responsive to context but substantively insensitive to most of it. Heuristic collapse is distinct from factual error or generic advice — a model exhibiting it may produce fluent, plausible, individually coherent recommendations while still failing to integrate the full context it was given. This makes it difficult to detect from spot-checking individual outputs, and easy to miss in standard benchmarks that evaluate accuracy rather than input sensitivity. This behavior is consistent with the broader phenomenon of shortcut learning \citep{geirhos2020shortcut}, in which models adopt decision rules that track salient surface features rather than performing the deeper reasoning the task requires.

We study heuristic collapse in the domain of investment advice, where investment suitability standards provide a principled and legally grounded account of what genuine personalization requires. A suitable recommendation must reflect a client's risk tolerance, age, income, investment horizon, and liquidity needs rather than any one of these in isolation. A recommendation that collapses this multi-factor assessment to a single dominant input is not merely incomplete --- it is unsuitable by the legal standard that governs financial advice.

To detect and quantify heuristic collapse in investment advice, we construct a diagnostic pipeline: we generate 1,000 synthetic client profiles via Latin hypercube sampling to ensure broad coverage of the input space, collect investment recommendations from frontier LLMs, fit interpretable surrogate models to the resulting input-output pairs, and measure the feature concentration of the fitted surrogates as a proxy for heuristic collapse. Applying this pipeline, we find that allocation decisions across models are dominated by risk tolerance, with other suitability-relevant features contributing marginally. We further find that web search partially attenuates heuristic collapse, but the effect is uneven.

\section{Approach}
\label{approach}

We design a controlled study to evaluate whether frontier LLMs exhibit heuristic collapse when acting as investment advisors. We generate synthetic client profiles using Latin hypercube sampling to ensure broad, near-orthogonal coverage of the input space, then query frontier LLMs to construct portfolios from a standardized set of 20 investment products. To measure heuristic collapse, we fit interpretable surrogate models to the resulting input-output pairs and diagnose feature concentration to measure how much the model's decisions depend on any one feature within the full context it is given. We additionally assess the effect of web search on heuristic collapse to reflect realistic deployment scenarios.

Our findings do not presuppose that there exists a single correct portfolio allocation for any given client profile. Investment suitability is not a function with a known closed-form solution: reasonable advisors will disagree on the precise allocations appropriate for a given client, and optimal portfolios depend on assumptions about expected returns, correlations, and risk aversion that are themselves contested. Our claim is more modest and, we argue, more robust: \textit{whatever} the correct mapping from client profile to portfolio, it should not be well-approximated by a single input feature. A decision rule that concentrates on self-reported risk tolerance while remaining largely insensitive to age, income, investment horizon, and other relevant factors is inadequate not because it disagrees with some benchmark allocation, but because the structure of the decision function itself is inconsistent with what suitability requires. The absence of ground truth does not preclude diagnosing heuristic collapse; it only requires shifting the criterion from output accuracy to input sensitivity.

\subsection{Generating inputs with Latin Hypercube Sampling}

Similar to ablation studies in machine learning, we systematically vary client characteristics while holding others constant. To achieve this, we use Latin hypercube sampling, a quasi-random sampling method that ensures (1) good coverage across the range of each characteristic and (2) low correlation between characteristics. This approach is standard in sensitivity analysis and experimental design when the goal is to isolate causal effects. The resulting profiles span realistic ranges but with orthogonal variation: the correlations between age and income, age and savings, and income and savings are all less than 0.07. This orthogonality enables regression analysis to identify which characteristics independently predict LLM recommendations. See Appendix \ref{app:client_profiles} for additional details about the client profiles. 

\subsection{Fitting surrogate models}

Given the LLM's input-output pairs, we fit a surrogate model to approximate the LLM's decision function. Let $\mathbf{x} \in \mathcal{X}$ be the input profile and $y_d \in \mathbb{R}$ denote the LLM's allocation output for asset class $d$. We observe $n$ input-output pairs $\{(\mathbf{x}_i, y_{d,i})\}_{i=1}^{n}$ collected by querying the LLM across our synthetic client profiles. For each asset class $d$, we fit a surrogate model $\hat{f}_d : \mathcal{X} \rightarrow \mathbb{R}$ by selecting the best model from a candidate set $\mathcal{M} = \{\text{Random Forest, Ridge Regression}\}$ via grid search over hyperparameters with 5-fold cross-validated $R^2$:
\begin{equation}
    \hat{f}_d = \operatorname*{arg\,max}_{m \in \mathcal{M},\, \boldsymbol{\theta}} \; R^2_{\mathrm{CV}}\!\left(m_{\boldsymbol{\theta}},\, d\right)
\end{equation}

Input features are preprocessed automatically: numeric columns are standardized and categorical columns are one-hot encoded. Each tree is evaluated via 5-fold cross-validated $R^2$ to avoid overfitting, and refit on the full dataset to obtain stable importance estimates.

\subsection{Measuring heuristic collapse}

When the selected surrogate is a random forest, each $\hat{f}_d$ yields a feature importance vector $\mathbf{w}_d \in \mathbb{R}^k_{\geq 0}$, where $k$ is the number of input features after preprocessing and $w_{d,j}$ is the mean decrease in impurity attributable to feature $j$ across all trees. When Ridge Regression is selected, $w_{d,j} = |\beta_j|$, the magnitude of the standardized coefficient for feature $j$. In both cases, we normalize importances so they sum to one:

\begin{equation}
    s_{d,j} = \frac{w_{d,j}}{\displaystyle\sum_{\ell=1}^{k} w_{d,\ell}}
\end{equation}

To measure heuristic collapse for output dimension $d$, we adapt the Herfindahl--Hirschman Index \citep{rhoades1993herfindahl} to measure feature concentration (FC):

\begin{equation}
    \text{FC}_d = \sum_{j=1}^{k} s_{d,j}^2
\end{equation}

$\text{FC}_d$ ranges from $\frac{1}{k}$ (importance distributed uniformly across all features --- holistic integration) to $1$ (all importance concentrated in a single feature --- pure heuristic).

\section{Results}
\label{results}

We prompt LLMs to act as financial advisors by providing a detailed client profile and a menu of investment products. The model must generate a comprehensive investment recommendation that includes: (1) product selection from the provided options, (2) specific allocation percentages that sum to 100\%, and (3) rationale behind its recommendation. We use structured outputs to ensure recommendations follow a consistent format and include all required components. This task structure mirrors real-world advisory interactions in which advisors must synthesize client information, evaluate available investment vehicles, and construct personalized portfolios. By standardizing the product menu across all profiles, we isolate the model's ability to provide suitable advice based on client characteristics rather than confounding the selection with product knowledge variability. 

We evaluate GPT family of models --- GPT-4o \citep{hurst2024gpt}, GPT 5.4 Nano, GPT 5.4 Mini, and GPT 5.4 --- using 1,000 synthetic client profiles. We focus on the GPT model family to hold training paradigm and tool-calling implementation constant, treating model capability tier as the primary axis of variation. Prior work evaluates how standalone LLMs give investment advice, but deployed LLM advisors will likely be agentic systems with web search. We therefore evaluate each model under two conditions: a baseline condition with no tools, and a web search condition in which the model must use web search before generating a recommendation.

\subsection{What LLMs Recommend}

We measure portfolio construction along two dimensions: diversification and personalization. For diversification, we use the Herfindahl--Hirschman Index in its canonical formulation (HHI, Equation~\ref{eq:hhi}), where lower values indicate greater spread across asset classes. Let $\mathcal{C}$ denote the set of asset classes and let $w_{p,c}$ denote the total portfolio weight allocated to asset class $c$ in portfolio $p$, where $\sum_{c \in \mathcal{C}} w_{p,c} = 1$:

\begin{equation}
    \text{HHI}(p) = \sum_{c \in \mathcal{C}} w_{p,c}^2
    \label{eq:hhi}
\end{equation}

For personalization, we use weighted Jaccard similarity (Equation~\ref{eq:jaccard}), where higher values indicate greater cross-client similarity and therefore less personalization. Let $\mathcal{A}$ denote the universe of assets, and let $w_{p,i}$ and $w_{q,i}$ denote the weights of asset $i$ in portfolios $p$ and $q$ respectively (with $w = 0$ for assets not held):

\begin{equation}
    J(p, q) = \frac{\sum_{i \in \mathcal{A}} \min(w_{p,i},\, w_{q,i})}
                   {\sum_{i \in \mathcal{A}} \max(w_{p,i},\, w_{q,i})}
    \label{eq:jaccard}
\end{equation}

\begin{table}[t]
\centering
\begin{tabular}{lcccc}
\toprule
& \multicolumn{2}{c}{\textbf{HHI (Diversification)}} & \multicolumn{2}{c}{\textbf{Jaccard (Similarity)}} \\
\cmidrule(lr){2-3} \cmidrule(lr){4-5}
\textbf{Model} & Baseline & $\Delta_{\text{Search}}$ & Baseline & $\Delta_{\text{Search}}$ \\
\midrule
GPT-4o       & $0.19 \pm 0.04$ & $\bm{-0.06}^{***}$ & $0.26 \pm 0.05$ & $+0.00^{***}$ \\
GPT-5.4 Nano & $0.19 \pm 0.05$ & $\bm{-0.04}^{***}$ & $0.33 \pm 0.07$ & $\bm{+0.09}^{***}$ \\
GPT-5.4 Mini & $0.20 \pm 0.05$ & $\bm{-0.05}^{***}$ & $0.38 \pm 0.08$ & $\bm{+0.04}^{***}$ \\
GPT-5.4      & $0.17 \pm 0.06$ & $\bm{-0.01}^{***}$ & $0.36 \pm 0.06$ & $\bm{+0.07}^{***}$ \\
\bottomrule
\end{tabular}
\caption{Effect of web search on portfolio diversification (HHI) and personalization (Jaccard similarity). HHI ranges from 0.05 (uniform allocation across all 20 products) to 1.0 (fully concentrated); lower is better. Jaccard similarity is computed pairwise across all client pairs and averaged; lower values indicate more client-specific recommendations. Baseline = No tools condition. $\Delta_{\text{Search}}$ = Search $-$ No Tools. Bold indicates a practically meaningful deterioration ($d > 0.2$) or improvement ($d < -0.2$). Significance: $^{***}p < 0.001$ via paired Wilcoxon signed-rank test.}
\label{tab:portfolio_metrics}
\end{table}

\textbf{LLMs produce moderately diversified but homogeneous portfolios.} Suitable advice should produce both low HHI and low Jaccard similarity. As shown in Table~\ref{tab:portfolio_metrics}, LLMs without web search produce moderately diversified portfolios well below the maximum concentration score of 1.0 but above the theoretical minimum of 0.05 for a uniformly distributed portfolio. Personalization varies more substantially across models: GPT-4o produces the most client-specific recommendations, while the GPT-5.4 family clusters higher, which indicates greater cross-client homogeneity.

\textbf{Web search improves diversification but reduces personalization.} We find that web search improves diversification across all four models ($p < 0.001$). Three of the four models show practically meaningful effects ($|d| > 0.2$): GPT-4o exhibits the largest improvement and GPT-5.4 shows the smallest HHI reduction ($\Delta = -0.01$). However, web search simultaneously increases cross-client similarity for three of the four models ($|d| = 0.55$--$1.76$). The sole exception is GPT-4o, whose Jaccard similarity is essentially unchanged by web search ($|d| = 0.02$) and is the only model for which tools improve diversification without meaningfully homogenizing recommendations. For the GPT-5.4 family, tools appear to anchor recommendations toward a common set of current products, reducing the client-specific tailoring that characterizes baseline behavior. This trade-off --- improved diversification at the cost of personalization --- raises questions about whether tool-augmented recommendations better satisfy fiduciary suitability standards, since suitable advice requires both diversification and personalization.

\subsection{Heuristic Collapse}

These metrics characterize the outputs of LLM recommendations, but do not reveal what drove them. To identify which client characteristics drive financial advice recommendations, we reverse-engineered LLM decision-making using supervised machine learning. For each asset class, we fit surrogate models to predict allocation percentages from client profile factors (age, income, risk tolerance, investment timeline, etc.). We then measured both predictive accuracy ($R^2$) and feature concentration (FC) of the fitted models. Surrogate fidelity is a precondition for interpreting FC: when $R^2$ is low, the surrogate has failed to capture the LLM's decision function, and concentration estimates are unreliable. We therefore report $R^2$ alongside FC throughout and caution against drawing strong conclusions from FC values where surrogate fit is poor. High $R^2$ paired with high FC is the diagnostic signature of heuristic collapse; high $R^2$ paired with low FC indicates holistic integration; and low $R^2$ in either case indicates that the LLM's allocation behavior for that asset class is not well-characterized by a simple mapping from the input features we measured.

\begin{figure}[t]
    \centering
    \vspace{-1em}
    \includegraphics[width=\linewidth]{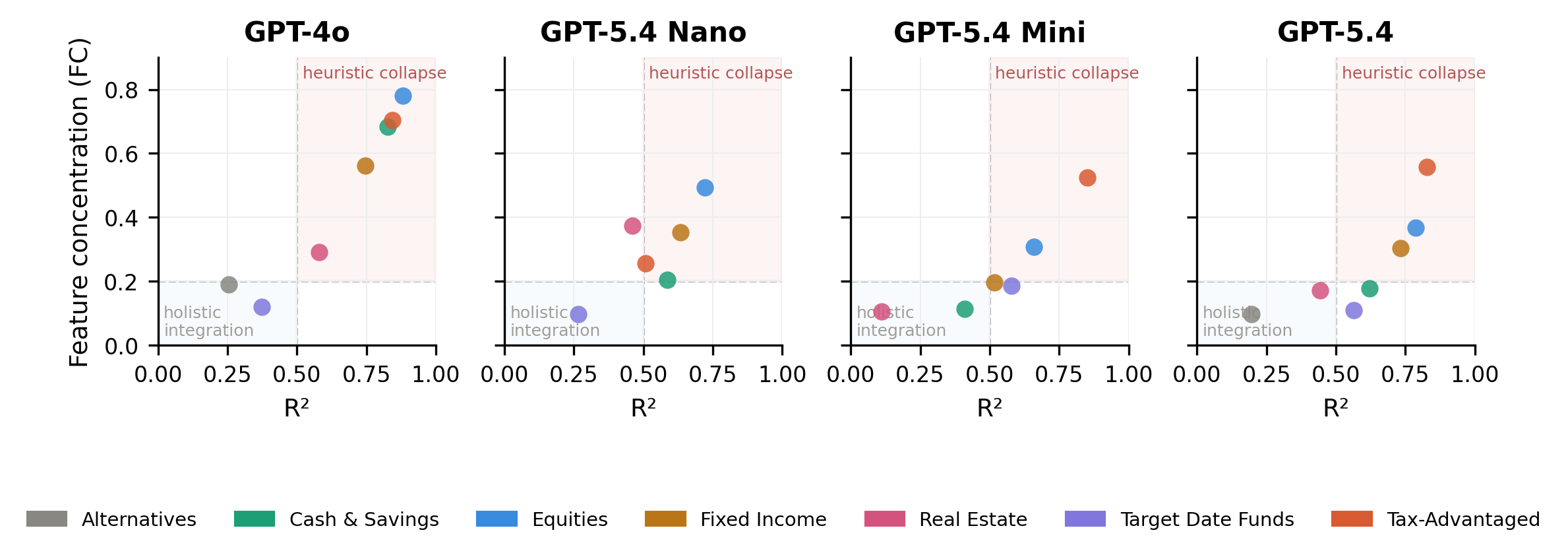}
    \caption{Comparison of reverse-engineered surrogate regressors for LLM allocation decisions without web search. $R^2$ measures predictive performance and FC measures feature concentration. 20 investment products have been aggregated into 7 asset classes for ease of reporting.}
    \label{fig:heuristics}
\end{figure}

\textbf{LLM allocation decisions are dominated by heuristics.} As shown in Figure~\ref{fig:heuristics}, LLMs exhibit substantial heuristic collapse across most asset classes, though the degree varies by model and asset class. For GPT-4o, equities (FC = 0.780, $R^2 = 0.882$) and tax-advantaged accounts (FC = 0.704, $R^2 = 0.843$) show the highest heuristic use. This indicates that a small number of client features --- most prominently self-reported risk tolerance --- dominate allocation decisions. Cash and savings similarly shows high heuristic use (FC = 0.683, $R^2 = 0.827$). The GPT-5.4 family exhibits less heuristic collapse overall. Across most models, recommendations follow a simple heuristic: aggressive clients receive equities, conservative clients receive bonds, with self-reported risk tolerance accounting for 57–88\% of predictive weight  and minimal integration of income. Age and investment timeline appear as secondary factors but remain substantially outweighed, except in GPT-5.4 where feature use is more distributed. The reliance of recommendations on self-reported risk tolerance is consistent with findings in \citet{fieberg2025using}. See Appendix \ref{app:surr_features} for details. 

\textbf{Web search partially attenuates heuristic collapse, but the effect is uneven.} Web search partially mitigates heuristic collapse, but the effect is uneven across asset classes and models. For GPT-4o, web search produces the largest reductions in heuristic use for cash and savings ($-65\%$ FC) and tax-advantaged accounts ($-45\%$ FC), while equities (+16\% FC) and fixed income (near-zero) are more resistant to change. Predictive accuracy declines alongside concentration for most asset classes, which indicates that recommendations generated with web search draw on a broader and less redundant set of client characteristics. For the GPT-5.4 family, the effects are more muted: heuristic use is already lower, and web search produces smaller absolute reductions --- and in some cases marginal increases, as seen in equities for GPT-5.4 Mini and Nano. 

\subsection{Qualitative Analysis}

To complement the quantitative portfolio metrics, we analyze the rationale text produced by each model with and without web search. For a fixed random sample of $n = 100$ clients per model we use an LLM-as-judge evaluation   \citep{zheng2023judging}, in which judges from two other model families (Claude Sonnet 4.6, Gemini 2.5 Pro) express a pairwise preference over each rationale pair on four dimensions: market grounding (whether the rationale cites current rates or economic conditions), specificity (whether justifications reference concrete numbers, named products, or explicit client attributes), profile relevance (how well the reasoning is tailored to the individual client's profile), and depth (overall completeness and thoroughness of the reasoning). Inter-rater agreement between the two judges is moderate across all models evaluated (mean $\kappa$ = 0.44–0.58).

\textbf{Web search fails to improve rationale quality for some LLMs but improves for others.} The LLM judge results reveal a divergence across models (Table ~\ref{tab:search_win_rate}). GPT-4o is the sole model for which web search fails to improve rationale quality across all four dimensions: win rates fall well below 0.5 for market grounding ($0.10 \pm 0.02$), specificity ($0.19 \pm 0.08$), profile relevance ($0.17 \pm 0.02$), and depth ($0.23 \pm 0.09$). GPT-5.4 Nano and GPT-5.4 Mini demonstrate improvements along certain dimensions, but the improvements are not consistent across all four dimensions. By contrast, GPT-5.4 shows consistent and substantial improvements across all four dimensions.

\textbf{This divergence reflects whether web search displaces or augments parametric reasoning.} For GPT-4o, rationales generated without web search more frequently name concrete investments and link allocations to specific client attributes, whereas web search rationales tend to be shorter, more abstract, and occasionally self-contradictory; one judge observes that a required rationale ``pivots to `safety and low-risk' without any data, numbers, or named products to support either framing'' despite the client having an aggressive risk profile. For the GPT-5.4 family, the pattern reverses: judges highlight that web search rationales introduce current rate conditions, cite concrete allocation figures, and engage more explicitly with individual client attributes such as debt load, liquidity, and tax situation. We hypothesize that GPT-4o's parametric reasoning is displaced by the web-search requirement, whereas the GPT-5.4 family is able to integrate retrieved information without sacrificing --- and indeed improving --- the quality of its justifications. However, neither model-tool combination simultaneously satisfies all dimensions of fiduciary suitability, which requires not only appropriate risk exposure but individualized reasoning.

\begin{table}[t]
\centering
\small
\begin{tabular}{lcccccccc}
\toprule
& \multicolumn{2}{c}{GPT-4o} 
& \multicolumn{2}{c}{GPT-5.4 Nano} 
& \multicolumn{2}{c}{GPT-5.4 Mini} 
& \multicolumn{2}{c}{GPT-5.4} \\
\cmidrule(lr){2-3}\cmidrule(lr){4-5}\cmidrule(lr){6-7}\cmidrule(lr){8-9}
Dimension & Win & Tie & Win & Tie & Win & Tie & Win & Tie \\
\midrule
Market grounding  & $\mathbf{0.10}$ & $0.90$ & $\mathbf{0.45}$ & $0.55$ & $0.96$ & $0.05$ & $1.00$ & $0.00$ \\
Specificity       & $\mathbf{0.19}$ & $0.30$ & $0.85$ & $0.07$ & $0.63$ & $0.24$ & $0.72$ & $0.16$ \\
Profile relevance & $\mathbf{0.17}$ & $0.21$ & $0.87$ & $0.10$ & $\mathbf{0.45}$ & $0.49$ & $\mathbf{0.37}$ & $0.60$ \\
Depth             & $\mathbf{0.23}$ & $0.15$ & $0.86$ & $0.08$ & $0.85$ & $0.10$ & $0.83$ & $0.14$ \\
\bottomrule
\end{tabular}
\caption{Pairwise judge preferences for rationales generated with vs.\ without web search. Win = judges prefer web search rationale; Tie = no preference; Loss = $1 - \text{Win} - \text{Tie}$. \textbf{Bold} indicates win rates below 0.5.}
\label{tab:search_win_rate}
\end{table}

\section{Related Work}
\label{sec:related_work}

The question of whether LLMs can provide genuine personalized advice in high-stakes domains has received growing attention, but existing evaluations predominantly assess output quality. A parallel literature has established that LLMs exhibit systematic cognitive biases such as anchoring, framing, and shortcut reasoning. This literature suggests that LLMs can produce plausible outputs while masking flawed decision processes. Whether these tendencies compromise personalization in advisory settings has not been definitively examined. We draw on three bodies of work to address this gap: interpretability methods for attributing model decisions, the literature on shortcut learning and cognitive bias in LLMs, and research on automated financial advice as our empirical domain.

\textbf{Surrogate models and interpretability.} The surrogate model literature is mature for traditional ML but has not been applied to reverse-engineer LLM decision-making at the population level. LIME \citep{ribeiro2016should} and SHAP \citep{lundberg2017unified} are the foundational local explanation methods that explain individual predictions by fitting local approximations around a single input. For LLMs, mechanistic interpretability methods such as probing classifiers \citep{belinkov2022probing} and activation patching \citep{meng2022locating} have made progress in identifying where factual associations are stored inside transformer weights, but these techniques operate at the level of internal representations rather than observable input--output behavior. Closer to our approach, \citet{hernandez2023inspecting} and related work on causal tracing characterize how specific input features influence model outputs, but do so one example at a time. Behavioral auditing frameworks such as CheckList \citep{ribeiro2020beyond} and HELM \citep{liang2022holistic} evaluate LLM outputs across structured input populations, but are designed to surface failures rather than to quantify which input dimensions drive output variation. Our method bridges these two traditions: we use a designed synthetic population of inputs and fit a surrogate model to the resulting input--output distribution, which enables population-level attribution of LLM decision-making without access to model internals.

\textbf{Shortcut learning and heuristic collapse.} A growing body of work establishes that LLMs exhibit systematic cognitive biases inherited from human-like heuristic reasoning. \citet{suri2024large} demonstrate that GPT 3.5 exhibits classic biases from the psychological literature such as anchoring, framing, and availability. Subsequent work has found that behavioral biases is pervasive across model families and persists under standard debiasing strategies such as chain-of-thought prompting \citep{ ross2024llm, lou2026anchoring}. \citet{echterhoff2024cognitive} test 13,465 prompts for anchoring, framing, status quo, and primacy biases in high-stakes decision contexts and find these biases present across both commercial and open-source LLMs. Critically, \citet{malberg2025comprehensive} evaluate 30 cognitive biases across 20 models and find no clear relationship between model scale and overall bias levels, which suggests that scaling alone does not reduce heuristic reliance.

\textbf{Robo-advisors, suitability, and financial LLMs.} Robo-advisors are algorithmic systems that provide automated and affordable financial advice. Because they offer financial advice, robo-advisors are held to a fiduciary standard \citep{fein2017robo}. A core component of fiduciary duty is the suitability requirement. Legal scholars and regulators have questioned whether automated systems can satisfy this personalization obligation, as suitability demands an individualized assessment rather than ``one size fits all'' advice \citep{sec2017robo}. \cite{lo2024can} argue that LLMs have potential as financial advisors but require fiduciary guardrails. \cite{fieberg2025using} test 32 LLMs on 64 investor profiles and find that LLMs respond appropriately to stated risk tolerance. \cite{takayanagi2025generative} examine interactive LLM-advisors and find they match human performance in eliciting client preferences but demonstrate failure modes when elicitation is inaccurate. We extend this line of work by examining agentic LLMs with web search --- a more realistic deployment scenario --- and use reverse-engineering to reveal how LLMs make allocation decisions.

\section{Discussion}
\label{sec:discussion}

Our findings reveal a consistent pattern: LLMs acting as financial advisors exhibit heuristic collapse, compressing multi-factor suitability determinations into decisions driven primarily by a single input dimension while remaining largely insensitive to the broader client context. It is not yet clear if this is a domain-specific failure, and in domains without such clear fiduciary standards like finance --- medical triage, legal interpretation, educational advising --- the same failure mode may be harder to detect but no less consequential.

\textbf{Suitability requires capacity and willingness.} Our finding that LLM allocations are dominated by self-reported risk tolerance is consistent with \citet{fieberg2025using}, who document the same pattern across a broader set of LLMs and treat it as broadly appropriate given the tenants of modern portfolio theory. However, self-reported risk tolerance is itself a noisy proxy for true risk preferences \citep{holt2002risk}. Investors systematically misreport risk tolerance --- overestimating it in bull markets, underestimating it under recent losses --- and preferences elicited through a single questionnaire item are known to have low test-retest reliability \citep{pan2012questionnaires}. Advice that faithfully mirrors a stated preference may therefore be unsuitable even when it appears well-calibrated.

More fundamentally, financial suitability standards require advisors to assess both \textit{willingness} to take risk (risk tolerance, investment objectives) and \textit{capacity} for risk (financial resources, time horizon, liquidity needs). LLMs focus almost entirely on willingness, which produces recommendations that may be unsuitable even when they match stated risk preferences. By failing to integrate age, income, investment timeline, and liquidity into holistic assessments, LLMs generate advice that could expose clients to inappropriate risk levels regardless of their stated tolerance. Web search partially mitigates this by incorporating more client characteristics into its recommendation, but gaps remain.

\textbf{Future directions.} This work suggests that current LLMs are likely not suitable financial advisors. Future research should examine: (1) how different model families (open-source vs. closed-source, different model sizes) affect advice suitability; (2) whether additional tool use and fine-tuning improves holistic reasoning; (3) whether multi-turn conversations where LLMs actively gather information produce better outcomes than single-shot recommendations; and (4) validation on realistic client profiles sampled from actual financial data (e.g., Survey of Consumer Finances). 

\section{Conclusion}
\label{sec:conclusion}

We set out to ask whether frontier LLMs provide genuinely personalized financial advice, or merely the appearance of it. The evidence points consistently toward the latter. We find that LLM investment recommendations exhibit systematic heuristic collapse: allocation decisions are dominated by a client's self-reported risk tolerance, with age, income, investment horizon, and liquidity needs contributing minimally despite being central to fiduciary suitability standards. This failure is not visible in individual outputs; it is a population-level failure of input sensitivity that standard spot-checking cannot detect.

Web search complicates this picture without resolving it. Web search improves portfolio diversification across all models and reduces heuristic collapse for several asset classes. However, the effect is uneven: recommendations produced with web search are more homogeneous across clients and do not necessarily produce better rationales behind investment recommendations. We hypothesize that web search sometimes displaces the model's parametric reasoning rather than supplementing it, which results in portfolios that look more diversified while being less suited to the individual. 

Together, these findings suggest that adding web search or scaling model capability may not be sufficient for genuinely personalized financial advice. What matters is not just output quality but input sensitivity --- whether the model actually integrates a client's full profile rather than anchoring on stated risk tolerance alone. Until that can be demonstrated, LLMs are better understood as generators of plausible-sounding advice than as sources of suitable recommendations.

\section*{Acknowledgements}
We thank Allen Ferrell and Thomas J. Brennan for their helpful discussions.

\bibliography{iclr2026_conference}
\bibliographystyle{iclr2026_conference}

\newpage

\appendix
\section{Appendix}

\subsection{Characteristics of Synthetic Client Profiles}
\label{app:client_profiles}

\begin{table}[!h]
\centering
\begin{tabular}{lllrl}
\toprule
Dimension & Type & Range/Levels & Unique Values & Coverage \\
\midrule
Age & Continuous & 25 - 75 years & 11 & Full range \\
Annual Income & Continuous & $15,000 - $920,000 & 8 & Full range \\
Total Savings & Continuous & $0 - $8,192,000 & 15 & Full range \\
Outstanding Debt & Continuous & $0 - $2,048,000 & 13 & Full range \\
Dependents & Continuous & 0 - 5 & 6 & Full range \\
Risk Tolerance & Categorical & 4 levels & 4 & Balanced \\
Investment Experience & Categorical & 5 levels & 5 & Balanced \\
Investment Timeline & Categorical & 4 levels & 4 & Balanced \\
Education & Categorical & 5 levels & 5 & Balanced \\
Marital Status & Categorical & 4 levels & 4 & Balanced \\
\bottomrule
\end{tabular}
\caption{Dimensions and coverage of variables in synthetic client profiles.}
\label{tab:dimension_coverage}
\end{table}

\begin{table}[!h]
\centering
\begin{tabular}{lr}
\toprule
Variable & Categories \\
\midrule
Risk Tolerance & Conservative, Moderate, Moderately Aggressive, Aggressive \\
Investment Experience & None, Limited, Some, Moderate, Extensive \\
Investment Timeline & 1-5, 5-15, 15-30, 30+ Years \\
Education & High School, Associate's, Bachelor's, Master's, Doctorate Degree \\
Marital Status & Single, Married, Divorced, Widowed \\
\bottomrule
\end{tabular}
\caption{Description of categorical variables included in client profiles.}
\label{tab:categorical_distributions}
\end{table}

\begin{table}[!h]
\centering
\begin{tabular}{lrrrrr}
\toprule
 & Age & Annual Income & Total Savings & Debt & Dependents \\
\midrule
Age & 1.000 & 0.003 & 0.025 & -0.007 & -0.005 \\
Annual Income & 0.003 & 1.000 & -0.021 & 0.007 & 0.010 \\
Total Savings & 0.025 & -0.021 & 1.000 & 0.069 & 0.004 \\
Debt & -0.007 & 0.007 & 0.069 & 1.000 & -0.020 \\
Dependents & -0.005 & 0.010 & 0.004 & -0.020 & 1.000 \\
\bottomrule
\end{tabular}
\caption{Correlation matrix of key continuous variables across the generated profile dataset.}
\label{tab:correlation_matrix}
\end{table}

\newpage

\subsection{LLM Prompts}
\label{ssec:prompts}

\begin{tcblisting}{
  title=Prompt: Baseline,
  colback=gray!5,
  colframe=black,
  listing only,
  breakable
}
You are an experienced financial advisor. Select investment products from the provided list and allocate across them based on the client's profile. Allocations must sum to 100%.

Client profile:
{profile}

{products}

Provide a rationale explaining how the client's profile drove your allocation decisions.
\end{tcblisting}

\begin{tcblisting}{
  title=Prompt: Investment Products,
  colback=gray!5,
  colframe=black,
  listing only,
  breakable
}
Available Investment Products:
1. High-Yield Savings Account
2. Money Market Funds 
3. Short-Term Treasury Bonds 
4. Long-Term Treasury Bonds 
5. Investment-Grade Corporate Bonds 
6. High-Yield Corporate Bonds 
7. Municipal Bonds 
8. Index Funds - S&P 500 
9. Index Funds - Total Stock Market 
10. Index Funds - International Stocks
11. Index Funds - Small Cap Stocks 
12. Target-Date Retirement Funds
13. Real Estate Investment Trusts (REITs)
14. Dividend Growth Stocks 
15. Growth Stocks
16. Commodities/Gold 
17. Cryptocurrency 
18. 529 College Savings Plan 
19. Health Savings Account (HSA) 
20. Certificates of Deposit (CDs) 
\end{tcblisting}

\begin{tcblisting}{
  title=Prompt: Addendum for Tool Required Condition,
  colback=gray!5,
  colframe=black,
  listing only,
  breakable
}
You have access to a web_search tool to look up current financial information, market data, interest rates, and investment research. Use it to inform your recommendations with up-to-date information. You MUST use web_search to look up: (1) current interest rates, (2) recent market performance, and (3) current economic conditions before providing advice.
\end{tcblisting}

\subsection{Structured Output Schema}
\label{ssec:schema}

Structured outputs were enforced using a JSON schema:

\begin{enumerate}
    \item \texttt{recommended\_products} --- an array of selected products, each with a name, type, allocation percentage, and free-text rationale;
    \item \texttt{rationale} --- a free-text description of the rationale behind portfolio construction.
\end{enumerate}

\subsection{LLM Web Search}

When web search was merely available or encouraged, the model almost never used them (0\% and 0.7\% of profiles, respectively), but when explicitly required, it used them nearly every time (99.9\% of profiles).

\begin{figure}[h]
    \centering
    \includegraphics[width=0.9\linewidth]{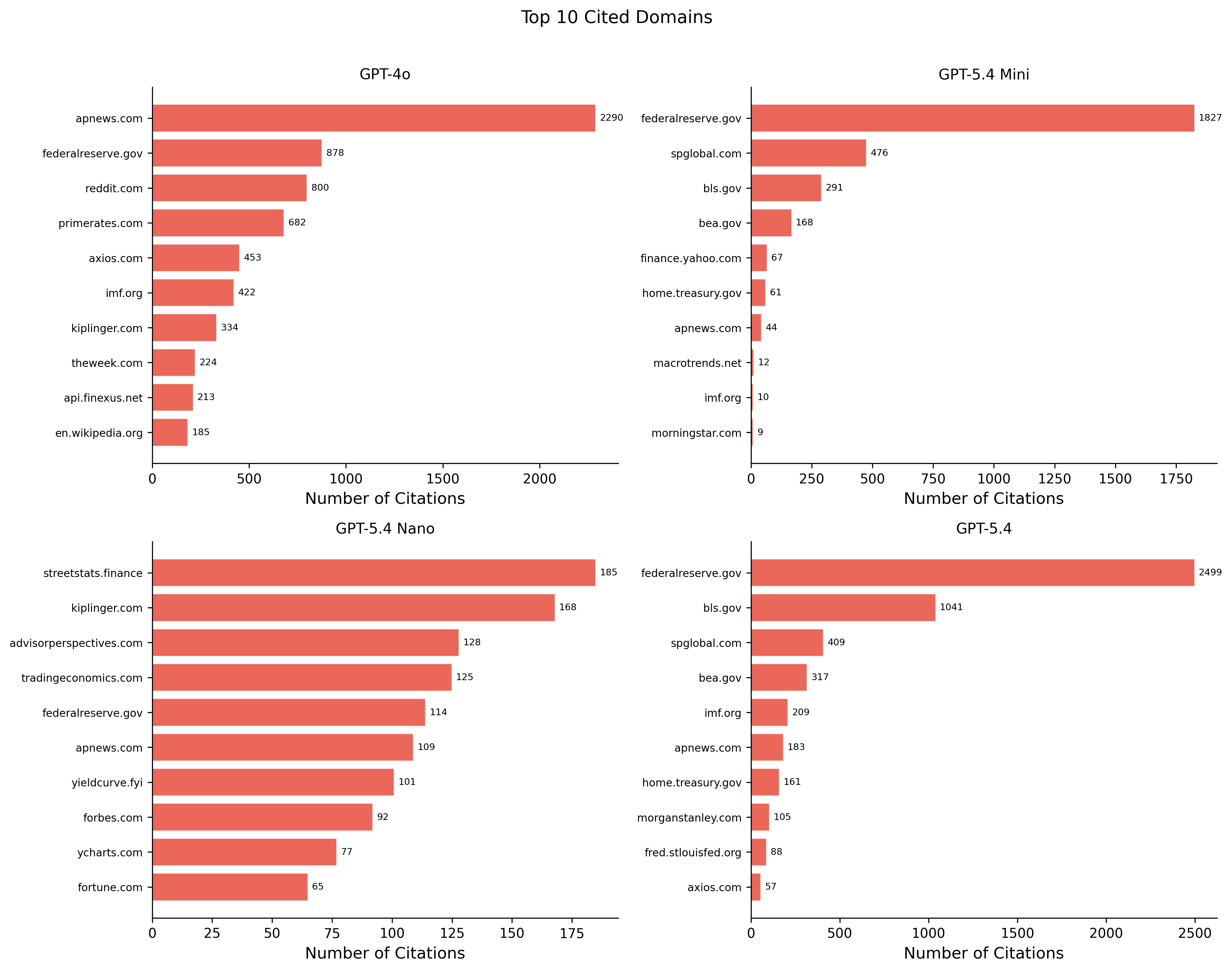}
    \caption{Citations generated by the LLM with required web search.}
    \label{fig:web_tools}
\end{figure}

\subsection{Round number heuristics} 

LLMs also exhibit a strong tendency to produce allocation percentages that are multiples of 5 (e.g.\ 10\%, 20\%, 25\%) rather than finer values --- a pattern that holds despite no such constraint being specified in the prompt. Without web search, between 89\% and 100\% of all allocation percentages are multiples of 5 across models, with GPT-4o being the most extreme (100\%). We quantify this tendency with a bias index $b = (\hat{p}_5 - 0.20) / (1 - 0.20)$, where $\hat{p}_5$ is the observed proportion of allocations divisible by 5 and 0.20 is the expected proportion under a uniform distribution; $b = 0$ indicates no excess rounding and $b = 1$ indicates every allocation is a multiple of 5. Baseline bias indices range from 0.87 to 1.00 across models.

Web search substantially reduces this bias for the GPT-5.4 family: the share of multiples-of-5 allocations falls from 89--97\% to 63--69\%, and bias indices drop from 0.87--0.96 to 0.54--0.61. The effect on GPT-4o is much smaller, with the bias index declining only marginally (1.00 $\to$ 0.95) and the multiple-of-5 share falling from 100\% to 96\%. These reductions are statistically significant across all models ($\chi^2$ tests, all $p < 0.001$; Mann--Whitney $U$ tests, all $p < 0.001$): while the shift is reliable, its practical magnitude varies considerably by model. The persistence of round-number bias even under required web search suggests that round-number anchoring is a deeply ingrained property of model behavior that web search alone does not fully override.

\subsection{Surrogate Features}
\label{app:surr_features}

\begin{table}[ht]
\centering
\small
\begin{tabular}{llrll r}
\toprule
& \multicolumn{2}{c}{No Web Search} & & \multicolumn{2}{c}{With Web Search} \\
\cmidrule(lr){2-3} \cmidrule(lr){5-6}
Asset Class & Top Feature & Share & & Top Feature & Share \\
\midrule
Equities            & Risk tolerance      & 88.2\% & & Risk tolerance      & 80.8\% \\
Tax-Adv.\ Accounts  & Has education goal  & 83.8\% & & Has education goal  & 60.7\% \\
Cash \& Savings     & Risk tolerance      & 82.4\% & & Risk tolerance      & 42.1\% \\
Fixed Income        & Risk tolerance      & 74.4\% & & Risk tolerance      & 75.0\% \\
Real Estate         & Risk tolerance      & 51.6\% & & Risk tolerance      & 49.3\% \\
Alternatives        & Risk tolerance      & 37.3\% & & Risk tolerance      & 28.1\% \\
Target-Date Funds   & Risk tolerance      & 26.1\% & & Risk tolerance      & 38.5\% \\
\bottomrule
\end{tabular}
\caption{Surrogate model results: GPT-4o}
\label{tab:surrogate_gpt4o}
\end{table}

\begin{table}[ht]
\centering
\small
\begin{tabular}{llrll r}
\toprule
& \multicolumn{2}{c}{No Web Search} & & \multicolumn{2}{c}{With Web Search} \\
\cmidrule(lr){2-3} \cmidrule(lr){5-6}
Asset Class & Top Feature & Share & & Top Feature & Share \\
\midrule
Tax-Adv.\ Accounts  & Has education goal  & 72.5\% & & Has education goal  & 61.4\% \\
Equities            & Risk tolerance      & 55.3\% & & Risk tolerance      & 51.1\% \\
Fixed Income        & Risk tolerance      & 51.9\% & & Risk tolerance      & 43.1\% \\
Cash \& Savings     & Liquid assets       & 29.8\% & & Liquid assets       & 47.7\% \\
Real Estate         & Risk tolerance      & 32.5\% & & Risk tolerance      & 40.8\% \\
Target-Date Funds   & Has retirement goal & 20.0\% & & Liquid assets       & 17.8\% \\
Alternatives        & Liquid assets       & 21.5\% & & Liquid assets       & 21.3\% \\
\bottomrule
\end{tabular}
\caption{Surrogate model results: GPT-5.4}
\label{tab:surrogate_gpt54}
\end{table}

\begin{table}[ht]
\centering
\small
\begin{tabular}{llrll r}
\toprule
& \multicolumn{2}{c}{No Web Search} & & \multicolumn{2}{c}{With Web Search} \\
\cmidrule(lr){2-3} \cmidrule(lr){5-6}
Asset Class & Top Feature & Share & & Top Feature & Share \\
\midrule
Tax-Adv.\ Accounts  & Has education goal  & 70.2\% & & Has education goal  & 51.3\% \\
Equities            & Risk tolerance      & 53.5\% & & Risk tolerance      & 62.5\% \\
Fixed Income        & Risk tolerance      & 38.9\% & & Risk tolerance      & 45.5\% \\
Target-Date Funds   & Has retirement goal & 37.9\% & & Has retirement goal & 24.1\% \\
Cash \& Savings     & Risk tolerance      & 22.9\% & & Risk tolerance      & 26.6\% \\
Real Estate         & Risk tolerance      & 21.9\% & & Risk tolerance      & 36.1\% \\
Alternatives        & ---                 & ---    & & Liquid assets       & 17.9\% \\
\bottomrule
\end{tabular}
\caption{Surrogate model results: GPT-5.4 Mini}
\label{tab:surrogate_gpt54mini}
\end{table}

\begin{table}[ht]
\centering
\small
\begin{tabular}{llrll r}
\toprule
& \multicolumn{2}{c}{No Web Search} & & \multicolumn{2}{c}{With Web Search} \\
\cmidrule(lr){2-3} \cmidrule(lr){5-6}
Asset Class & Top Feature & Share & & Top Feature & Share \\
\midrule
Equities            & Risk tolerance      & 68.9\% & & Risk tolerance      & 61.8\% \\
Real Estate         & Risk tolerance      & 60.0\% & & Risk tolerance      & 42.1\% \\
Fixed Income        & Risk tolerance      & 57.6\% & & Risk tolerance      & 54.4\% \\
Tax-Adv.\ Accounts  & Has education goal  & 44.7\% & & Has education goal  & 20.3\% \\
Cash \& Savings     & Risk tolerance      & 36.5\% & & Risk tolerance      & 31.7\% \\
Target-Date Funds   & Risk tolerance      & 17.5\% & & Savings rate        & 20.2\% \\
Alternatives        & ---                 & ---    & & Savings rate        & 15.6\% \\
\bottomrule
\end{tabular}
\caption{Surrogate model results: GPT-5.4 Nano}
\label{tab:surrogate_gpt54nano}
\end{table}

\end{document}